\documentclass[11pt]{article}

\usepackage{amsmath,amssymb,amsfonts,graphicx,amsthm,nicefrac,mathtools,bm}
\usepackage{hyperref}
\usepackage[T1]{fontenc}
\usepackage{bbm,mdframed}
\usepackage{enumerate,graphics,enumitem}
\usepackage[numbers]{natbib}
\usepackage[left=1in,right=1in,top=1in,bottom=1in]{geometry}
\usepackage{array}

\newtheorem{theorem}{Theorem}

\definecolor{verylightblue}{rgb}{0.7,0.8,1}
  {\begin{mdframed}[backgroundcolor=verylightblue]\begin{theorem}}%
  {\end{theorem}\end{mdframed}}

\definecolor{verylightgray}{gray}{0.95}
  {\begin{mdframed}[backgroundcolor=verylightgray]\begin{proof}}%
  {\end{proof}\end{mdframed}}

\newtheorem*{claim*}{Claim}
\newtheorem{claim}{Claim}
\newtheorem{thm}{Theorem}

\newtheorem{lemma}{Lemma}

\definecolor{verylightred}{rgb}{1,0.8,0.8}
  {\begin{mdframed}[backgroundcolor=verylightred]\begin{lemma}}%
  {\end{lemma}\end{mdframed}}

\newtheorem{proposition}{Proposition}
  {\begin{mdframed}[backgroundcolor=verylightblue]\begin{proposition}}%
  {\end{proposition}\end{mdframed}}

\theoremstyle{definition}
\newtheorem{definition}{Definition}
\theoremstyle{remark}

\makeatletter

\newtheorem*{rep@theorem}{\rep@title}
\newcommand{\newreptheorem}[2]
{\newenvironment{rep#1}[1]
{\def\rep@title{#2 \ref{##1}} \begin{rep@theorem}}%
 {\end{rep@theorem}}}
\makeatother
\newreptheorem{theorem}{Theorem}
\newreptheorem{lemma}{Lemma}
\newreptheorem{corollary}{Corollary}
\newreptheorem{proposition}{Proposition}

\newcommand{\thedate}{\today}
\newcommand{\theauthor}{Tijana Zrnic ~ ~ ~ Moritz Hardt\\
Department of Electrical Engineering and Computer Sciences\\
UC Berkeley\\
{\small \texttt{\{tijana.zrnic,hardt\}@berkeley.edu} }
}
\newcommand{\thetitle}{Natural Analysts in Adaptive Data Analysis}

\date{\thedate}
\author{\theauthor}
\title{\thetitle}

\usepackage[mmddyy]{datetime}

\usepackage{algorithm,algorithmic}

\makeatletter
\long\def\@makecaption#1#2{
        \vskip 0.8ex
        \setbox\@tempboxa\hbox{\small {\bf #1:} #2}
        \parindent 1.5em  
        \dimen0=\hsize
        \advance\dimen0 by -3em
        \ifdim \wd\@tempboxa >\dimen0
                \hbox to \hsize{
                        \parindent 0em
                        \hfil 
                        \parbox{\dimen0}{\def\baselinestretch{0.96}\small
                                {\bf #1.} #2
                                } 
                        \hfil}
        \else \hbox to \hsize{\hfil \box\@tempboxa \hfil}
        \fi
        }
\makeatother

\newcommand{\EE}{\mathbb{E}}
\newcommand{\D}{\mathcal{D}}
\newcommand{\M}{\mathcal{M}}
\newcommand{\A}{\mathcal{A}}

\newcommand{\X}{\mathcal{X}}
\newcommand{\Q}{\mathcal{Q}}

\newcommand{\F}{\mathcal{F}}
\newcommand{\Y}{\mathcal{Y}}

\newcommand{\HH}{\mathcal{H}}
\newcommand{\Ss}{\mathcal{S}}

\newcommand{\R}{\mathbb{R}}
\newcommand{\N}{\mathbb{N}}
\newcommand{\Oh}{\mathcal{O}}
\newcommand{\PP}{\mathbb{P}}
\newcommand{\Pp}{\mathcal{P}}

\date{}

\begin{document}

\maketitle

\begin{abstract}
Adaptive data analysis is frequently criticized for its pessimistic generalization guarantees. The source of these pessimistic bounds is a model that permits arbitrary, possibly adversarial analysts that optimally use information to bias results. While being a central issue in the field, still lacking are notions of natural analysts that allow for more optimistic bounds faithful to the reality that typical analysts aren't adversarial.

In this work, we propose notions of natural analysts that smoothly interpolate between the optimal non-adaptive bounds and the best-known adaptive generalization bounds. To accomplish this, we model the analyst's knowledge as evolving according to the rules of an unknown dynamical system that takes in revealed information and outputs new statistical queries to the data. This allows us to restrict the analyst through different natural control-theoretic notions. One such notion corresponds to a \emph{recency bias}, formalizing an inability to arbitrarily use distant information. Another complementary notion formalizes an \emph{anchoring bias}, a tendency to weight initial information more strongly. Both notions come with quantitative parameters that smoothly interpolate between the non-adaptive case and the fully adaptive case, allowing for a rich spectrum of intermediate analysts that are neither non-adaptive nor adversarial.

Natural not only from a cognitive perspective, we show that our notions also capture standard optimization methods, like gradient descent in various settings. This gives a new interpretation to the fact that gradient descent tends to overfit much less than its adaptive nature might suggest.
\end{abstract}

\section{Introduction}

Modern data analysis is usually adaptive in the sense that past analyses shape future analyses. This practice offers power and flexibility to data science at the cost of a greater potential for spurious results. The issue is now well recognized in multiple communities. The problem of \emph{inference after selection} is an active research area in statistics, while computer science has developed an area known as \emph{adaptive data analysis}.

The statistical community has focused on analyzing concrete two-step procedures, such as, variable selection followed by a significance test on the chosen variables~\cite{fithian2014optimal, belloni2014inference}. This approach leads to precise insight into some concrete procedures, but it does not capture the workflow of typical analysts that proceed in more than two steps.

Computer scientists took an alternative route by focusing on a powerful \emph{statistical query model} that in principle captures all sorts of different analyses involving many adaptive steps. In this model, an analyst interacts with a data set through a primitive called \emph{statistical queries}. In each round, the analyst can evaluate one statistical query on the data. Future statistical queries may depend arbitrarily on the revealed transcript of past queries and query results. This level of generality comes at the cost of diminished generalization ability. 

To review what's known, the generalization error on~$t$ non-adaptively chosen statistical queries on a data set of size~$n$ is on the order of $O(\sqrt{\log(t)/n}),$ as follows from Hoeffding's bound. In the fully adaptive model, Hoeffding's bound would only give a rate of~$\tilde O(\sqrt{t/n})$. This disappointing bound coincides with the naive strategy of splitting the data set into $t$ chunks, each of size $n/t$ and using one chunk for each query. Noise addition techniques combined with the mature technical repertoire of differential privacy yield a better bound of~$\tilde O( t^{1/4}/\sqrt{n} )$~\cite{bassily2016algorithmic}. However, this bound still features a polynomial dependence on the number of queries~$t$ that has resisted improvement for years. Negative results suggest that it might in fact be computationally hard to improve on this bound~\cite{hardt2014preventing,ullman2018limits}.

For years, the knee jerk reaction to such pessimistic bounds has been to point out that natural analysts aren't adversarial. However, it has proved challenging to formalize what makes natural analysts more benign than the worst-case bounds suggest. Indeed, to date there is still no comprehensive proposal for a class of analysts that allows for interesting intermediate points between the fully adaptive and non-adaptive case.

\subsection{Our Contributions}

In this work, we tackle the central conceptual challenge of formalizing classes of natural analysts using ideas from dynamical systems theory. Specifically, we model the analyst's knowledge as evolving according to the rules of an unknown dynamical system in discrete time. The system takes in query results~$a_t$ at each step and maintains a hidden state~$h_t$ at time~$t$. Based on its hidden state, the system chooses a new query $q_{t}=f_t(h_t)$ as a function of the hidden state (that may vary with time) and updates its hidden state $h_{t+1}=\psi_t(h_t, a_t)$ according to a state-transition map~$\psi_t$ that is allowed to vary with time. It is clear that we can recover the non-adaptive case by forcing the hidden state to be constant at all steps, whereas the fully adaptive case corresponds to an unrestricted hidden state and state transition rule. 

What is interesting is that this dynamical perspective allows us to restrict the analyst in natural ways, which we show lead to interesting trade-offs. These restrictions simultaneously correspond to natural control-theoretic notions, subsume common optimization procedures, and can be seen as formalizing well-known cognitive biases. We focus on two complementary notions of natural analysts that we call \emph{progressive} and \emph{conservative}.

\paragraph{Progressive analysts.}
Progressive analysts, intuitively speaking, have a recency bias and weight recent information more strongly than information received far into the past. We can think of a discount factor~$\lambda\in(0,1)$ by which the analyst downweights past observations. Formally, we call an analyst $\lambda$-\emph{progressive} if the state transition map is contractive\footnote{From here forward, we will assume $\|\cdot\|$ denotes some $\ell_p$-norm, $p\geq 1$.}: $\|\psi_t(h, a)-\psi_t(h', a)\|\le\lambda\|h-h'\|$.

To gain intuition, in the case of a linear state-transition map $h_{t+1}=Ah_t+Ba_t,$ this requirement corresponds to the condition~$\|A\|_{\text{op}}\le\lambda$, where $\|\cdot\|_{\text{op}}$ denotes the operator norm. In control-theoretic terms, this requirement expresses that the system is stable. Trajectories cannot blow up under repeated application of the state transition map. We show that this control-theoretic stability has a strong regularizing effect.

\begin{thm}[Informal result for progressive analysts]
There is a computationally efficient algorithm to answer $t$ statistical queries chosen adaptively by a $\lambda$-progressive analyst so that the error on each query is at most $\tilde O\left(\sqrt{K(\lambda)d_q\log(t)/n} \right)$, where $K(\lambda) = O\left(\frac{\log(1/(1-\lambda))}{\log(1/\lambda)}\right)$, $d_q$ is the dimension of the queries, and $n$ is the size of the data set.
\end{thm}

Since Theorem \ref{generalize1} allows queries of arbitrary dimension, $d_q$ can also be thought of as the number of parallel statistical queries in one round, making the total number of one-dimensional queries after $t$ rounds equal to $td_q$. With this in mind, we can see that for $\lambda=1-1/t,$ the bound reduces to the adaptive Hoeffding bound $\tilde O(\sqrt{t d_q/n})$ (by a first-order Taylor approximation). For any constant $\lambda$ bounded away from $1,$ we recover the non-adaptive bound. The proof of this result combines a simple compression argument with recent ideas in the context of recurrent neural networks~\cite{miller2018recurrent}.

We could hope that as $\lambda$ approaches~$1$ we not only recover the Hoeffding bound but rather the best known adaptive bounds that follow from differential privacy techniques. While this turns out to be difficult for progressive analysts for reasons we elaborate on later, we can indeed achieve this better trade-off for our second notion.

\paragraph{Conservative analysts.}
Conservative analysts favor initial information over new information in their decision making. Intuitively, this can be seen as a kind of anchoring bias. One of the ways we can express this is by requiring that the state-transition map gets increasingly Lipschitz in its second argument over time:
\[
\|\psi_t(h, a)-\psi_t(h, a')\|\le\eta\|\psi_{t-1}(h, a)-\psi_{t-1}(h, a')\|,
\]
for some $\eta\in(0,1)$. We call analysts satisfying this requirement $\eta^t$-\emph{conservative}, leading to the following result.
\begin{thm}[Informal result for conservative analysts, special case]
There is a computationally efficient algorithm to answer $t$ statistical queries chosen adaptively by a $\eta^t$-conservative analyst so that the error on each query is at most $\tilde O\left((K(\eta^t)d_q \log(t))^{1/4}/\sqrt{n} \right)$, where $K(\eta^t) = O\left(\frac{1}{\log(1/\eta)}\right)$, $d_q$ is the dimension of the queries, and $n$ is the size of the data set.
\end{thm}

Contrary to progressive analysts, if $\eta = 1 - 1/t$, the bound reduces to a multi-dimensional generalization of the hard-to-improve generalization bound $\tilde O( (td_q)^{1/4}/\sqrt{n})$.



\subsection{Proof Technique Overview}

The main technical tool used in our generalization proofs is an algorithmic abstraction called the \emph{truncated analyst}. For both progressive and conservative analysts, we design their respective truncated counterpart, which acts according to the same dynamics $\psi_t$. By construction, the truncated analyst has a time-independent number of rounds of adaptivity. We will also refer to the true analyst as the \emph{full analyst}, to contrast it with the corresponding truncated abstraction.

We first derive a natural conclusion stating that truncated analysts have time-independent generalization properties. Then, we show that, for a large enough level of truncation (which is still time-independent), the truncated analyst closely approximates the full one. This observation will enable us to claim that the full analyst, which is either progressive or conservative, inherits the generalization properties of its corresponding truncated version. One of the conclusions we will derive from here is the following: setting the parameters of progressiveness or conservatism to be constant with respect to the number of interactions yields a generalization error that scales only logarithmically with the number of queries.

\section{Analysts as Dynamical Systems}

\subsection{Problem Setting}

Let $\Ss:=\{X_1,\dots,X_n\}\in\D^n$ be a data set of $n$ i.i.d. samples from a
distribution $\Pp$ supported on $\D$. On one side, there is a data analyst, who
initially has no information about the drawn samples in $\Ss$. On the other side there is a statistical mechanism with access to $\Ss$, however with no knowledge of its true underlying distribution. At each time step $t\in\N$, the analyst and statistical mechanism have an interaction: the analyst asks a statistical query $q_t\in\Q$, and the statistical mechanism responds with an answer $a_t\in\A$. In the adaptive data analysis literature, statistical queries are typically defined as one-dimensional bounded functions, however in this work we generalize this definition to allow bounded functions in higher dimensions. The motivation for this is that many common procedures query a vector of values; for example, gradient descent queries a gradient of the loss at the current point. Formally, we define statistical queries as functions of the form $q_t:\D\rightarrow[0,1]^{d_q}$. In this generalized setting, a single query $q_t$ is equivalent to a set of $d_q$ one-dimensional queries. It is only natural to assume that the dimension of answers matches that of the posed queries, and hence we take $\A\subseteq\R^{d_q}$.

Before deciding on $q_t$, the analyst takes into account the previous interactions with the statistical mechanism, typically called the \emph{transcript}. In classical work on adaptive data analysis, the transcript at time $t$ consists of all query-answer pairs thus far, $(q_1,a_1,\dots,q_{t-1},a_{t-1})$. Recall that, in this work, the analyst only has access to the transcript through its hidden state, or history, $h_t\in\HH\subseteq \R^d$, acting according to the recursion:
\begin{equation}
\label{eq:analystdef}
    h_t = \psi_t(h_{t-1},a_{t-1}),
\end{equation}
where we initialize $h_0=0$. The variable $h_t$ serves as a possibly lossy encoding of the knowledge the analyst has gathered about data $\Ss$ up to time $t$. Based on this encoding, the analyst picks the next query $q_t\in\Q$:
\begin{equation}
\label{eq:query}
    q_t = f_t(h_t),
\end{equation}
where $f_t:\HH\rightarrow\Q$ is an arbitrary measurable function.

The goal of designing a statistical mechanism is to have the analyst learn about the \emph{distribution} $\Pp$, and not just the samples in $\Ss$. Mathematically, we want the \emph{generalization error} 
$$\max_{1\leq i\leq t}\|a_i - \EE_{X\sim\Pp}[q_i(X)]\|_\infty$$
to be small with high probability, for any given number of rounds $t$. The
difficulty is this task lies in the fact that the statistical mechanism
does not have access to $\Pp$. It might seem intuitive to set $a_t =
q_t(\Ss):=\frac{1}{n}\sum_{i=1}^n q_i(X_i).$ However, in general, this standard choice quickly leads to overfitting (see the paper \cite{blum2015ladder} for an example attack).

A better solution stems from a connection with privacy-preserving data analysis.
In particular, it has been shown that good \emph{sample accuracy} combined with
\emph{differential privacy} ensures small generalization error~\cite{dwork2015preserving, bassily2016algorithmic, dwork2015generalization}. 

We say that a possibly randomized function $\F:\D^n\rightarrow\Y\subseteq \R^d$ is $(\alpha,
\beta)$-differentially private for some $\alpha,\beta\geq 0$, if for all data sets $S,S'\in\D^n$, such that $S$ and $S'$ differ in at most one entry, it holds that:
$$\PP(\F(S)\in\Oh)\leq e^\alpha \PP(\F(S')\in\Oh) + \beta,$$
for any event $\Oh$. We will extensively rely on some of the well-known properties 
of differential privacy that we collect in the Appendix.

A possibly randomized function $\M:\D^n\times\Q\rightarrow\Y$, where
$\Y\subseteq\R^d$, is $(\epsilon, \delta)$-sample accurate if for every
data set $\Ss=\{X_1,\dots,X_n\}\in\D^n$ and every query $q\in\Q$, where
$q:\D\rightarrow\Y$, it holds that: $$\PP(\|\frac{1}{n}\sum_{i=1}^n q(X_i) -
\M(\Ss,q)\|_\infty\geq\epsilon)\leq \delta.$$

Applying these definitions to the problem of adaptive data analysis, we simultaneously want $\max_{1\leq i\leq t} \|a_i - q_i(\Ss)\|_\infty$ to be small, and $a_i$ to be constructed in a differentially private manner, thus obscuring the exact value of $q_i(\Ss)$. We will show that, with an appropriate choice of a differentially private mechanism, these two conditions will result in favorable generalization properties in our setting.

Our analysis will primarily make use of Gaussian noise addition,
as it achieves the hard-to-improve rate of $\tilde O(t^{1/4}/\sqrt{n})$, in the
one-dimensional statistical query model. We will use $\xi_t$ to denote a generic Gaussian noise vector; with this, the classical Gaussian mechanism is given by $a_t = q_t(\Ss) + \xi_t$, where $\xi_t$ is zero-mean noise of $d_q$ independent Gaussians with appropriately chosen variance.

The main idea for preventing adversarial behavior of analysts will be some form of contraction characterizing the state-transition map sequence $\{\psi_t\}$.  This approach
requires a way to translate closeness in norm into a form of information-theoretic closeness. In general, however, if two
different analysts have histories $h_t^1$ and $h_t^2$, such that $\|h_t^1 -
h_t^2\|\leq\epsilon$ for some very small $\epsilon>0$, it is impossible to say
whether their knowledge of $\Ss$ is indeed ``$\epsilon$-close''.  For this
reason, we introduce the assumption that $\HH$ is a discrete grid in $\R^d$ with
coordinate-wise resolution $\Delta>0$, where $\Delta$ is sufficiently small.
Mathematically, if $h=(h_1,\dots,h_d)\in\HH$, then $h_i = k_i\Delta$, for some
$k_i\in\mathbb{Z}$. This way, if two histories are close enough in norm, they
have to be semantically identical. This condition is satisfied by a great
majority of real-world data analysts. First, all ``transcripts'' generated by
numerical algorithms are memorized in computers using finite-bit precision.
Second, human analysts typically use only the first few digits after the decimal
of any performed numerical evaluation. It is also worth pointing out that all
generalization results obtained for the set $\HH$ also hold for all uniformly
discrete sets which have a packing radius at least $\Delta$.

\section{Progressive Analysts: Motivation and Generalization}
\label{progsec}

The first class of analysts is oblivious in that its knowledge of past
events diminishes over time.  We will aptly refer to such data analysts as
\emph{progressive}.

\begin{definition}[Progressive analyst]
\label{analyst1}
An adaptive analyst is \emph{$\lambda$-progressive} if the maps $\{\psi_t\}$ are $\lambda$-contractive in their first argument; for every $h,h'\in\HH$ and $a\in\A$, $\psi_t$ satisfies:
$$\|\psi_t(h,a) - \psi_t(h',a)\|\leq \lambda\|h - h'\|,$$
for some $\lambda\in [0,1]$. Additionally, we require $\psi_t(h,\cdot)$ to be $L$-Lipschitz for any fixed $h\in\HH$; that is, for all $a,a'\in\A$, ad some $L\geq 0$:
$$\|\psi_t(h,a)-\psi_t(h,a')\|\leq L \|a-a'\|.$$
\end{definition}

Without loss of generality, we also assume that the maps $\{\psi_t\}$ are
normalized to satisfy $\psi_t(0,0)=0$. This does not limit their expressiveness. 

We now motivate the definition of progressive analysts via three examples, before
proving our main generalization bound for this class of analysts.

It is well-known that humans exhibit numerous cognitive biases while performing
analytical tasks. One well-known bias is the recency
bias~\cite{cheadle2014adaptive}. This bias is defined as a tendency to focus
more on recent evidence than the history. 
%
We can think of recency bias as a motivating analogy for our definition of
progressive. In our definition, the parameter $\lambda$ determines how fast prior
information are forgotten.  The case $\lambda=0$ corresponds to full recency bias and
virtually no adaptivity in query formulation, while $\lambda=1$ implies no
recency bias and arbitrarily adaptive queries.

As another, contrasting example, iterative algorithms which interact with a fixed data set can also be thought of as adaptive analysts. Suppose that $\Ss$ contains simulation samples of an agent interacting with a stochastic environment, which returns noisy rewards from an unknown distribution and has known random transitions between a possibly large number of states. This problem can be modeled as a classical Markov decision process \cite{bertsekas2005dynamic}. Suppose that the analyst wishes to define a set of $d$ states, possibly by grouping the existing elementary states, such that the value function, which is the expected reward-to-go under the optimal policy, satisfies some criterion: for example, one objective could be maximizing the value function in one of the states of the model. First, the analyst initializes the set of states to some arbitrary set of fixed size $d$. Then, they recurse their hidden state, whose coordinates $i\leq d$ are updated as:
\begin{equation}
\label{eq:bellman}
    h_{t,i} = \sup_a (r_t(i,a) + \gamma\sum_{j=1}^d \PP(i,a,j)h_{t-1,j}),
\end{equation}
where the supremum is taken over the possible actions, $\gamma\in(0,1)$ is a discount factor, $\PP(i,a,j)$ is the probability of landing in state $j$ after taking action $a$ in state $i$, and $r_t(i,a)$ is the estimated average reward of taking action $a$ in state $i$. Equation \eqref{eq:bellman} is called the Bellman equation, and the algorithm given by repeated iterations of this equation is called value iteration \cite{bellman1957dynamic}, as it is used to find the value function. For example, if every sample $X_k\in\Ss$ is vector containing the initial state, action, reward, and subsequent state, $(s_{1,k}, a_k, r_k, s_{2,k})$, then the estimated reward is given by $r_t(i,a) = \sum_{k=1}^n r_k \mathbf{1}\{s_{1,k} = i, a_k = a\}/\sum_{k=1}^n \mathbf{1}\{s_{1,k} = i, a_k = a\}$. The analyst's queries are therefore asking for the reward estimates across all states and all actions. After running the Bellman update for a certain number of rounds, the analyst can now adaptively change the set of states, using the previously learned value of $h_t$ for initialization. Since the Bellman equation contracts $h_t$ by factor $\gamma$ in $\ell_\infty$-norm, such an analyst would be $\gamma$-progressive. The Bellman equation is at the core of numerous dynamic programs, thus making many algorithmic solvers of such problems progressive analysts.

Stable recurrent neural networks are another algorithmic example of progressive analysts.
Recurrent neural networks are given by the update: $$h_t = \rho(W h_{t-1} + U
a_{t-1}),$$ where $U\in\R^{d_q\times d}$, $W\in\R^{d\times d}$, and $\rho$ is a
point-wise non-linearity. The variable $a_t$ is the empirical answer to an
arbitrary query based on $h_t$. In this case, the analyst is
$\lambda$-progressive if $\|W\|_{\text{op}}\leq 1/L_\rho$, where $L_\rho$ is the
Lipschitz constant of the map $\rho$. For a detailed treatment of this case,
see the paper~\cite{miller2018recurrent}. The work also shows how other stateful models,
such as LSTMs, can be made stable and how stable models perform well in
practice.

Now we argue that the parametrization of progressive analysts allows interpolation between that of non-adaptive analysts and fully adaptive analysts. Then, we move on to proving the generalization error in regimes between these two extremes.

First, consider $L=0$. In this case, $h_t$ has no sensitivity to the answers of the statistical mechanism, so queries are trivially non-adaptive.


On the other end, $\lambda=1$ allows full adaptivity, for any $L>0$. To see this, imagine that $h_t$ is an infinite-dimensional vector\footnote{This can be formalized in the framework of separable Hilbert spaces, however this example is only intended to be illustrative.}, where each coordinate is initially $0$, and coordinate-wise, $h_t$ can take values in $L\A$. At time $t$, simply set the coordinates $(t-1)d_q + 1$ through $t d_q$ of $h_t$ to $L a_{t-1}$. Since all queries are computed via a deterministic function of the current history, which is composed by stacking the answers, the vector of all answers encodes the whole transcript in a lossless fashion. Consequently, this analyst is fully adaptive. One can easily verify that the described transition maps satisfy the conditions of Definition \ref{analyst1} with $\lambda=1$.

Since these two extreme cases reduce to generalization rates which are known from prior work, in the rest of this section we focus on the parameter set $\lambda\in[0,1)$, $L>0$.

\subsection{Truncated Analyst}

Now we introduce a useful counterpart of a $\lambda$-progressive analyst, who only has access to the last $k$ answers of the full analyst, for some constant $k$. This truncated analyst will be the main abstraction used in the proofs of this section.

Define the truncated analyst corresponding to a full progressive analyst as:
\begin{align*}
\label{eq:trunceqn1}
    h_t^k &= \psi_t(h_{t-1}^k,a_{t-1}), ~ ~ h_{t-j}^k = 0, ~ ~\forall j\geq k, \\
    q_t^k &= f_t(h_t^k),
\end{align*}
for fixed $k\in\N$. The truncated analyst updates their history according to the same map sequence as the full analyst, and receives exactly $k$ answers \emph{of the full analyst}.

First we show that, as aligned with intuition, each query of the truncated analyst has a time-independent generalization error.

\begin{lemma}
\label{boundedmem1}
Let $h_t^k$ be the history of a truncated analyst, and let the range of answers be of size $A^{d_q}$, where $A$ is polynomial in $n$. Then, at time $t$, the query $q_t^k$ asked by the truncated analysts satisfies the following:
$$\PP(\| q^k_t(\Ss) - \EE_{X\sim\Pp}[q^k_t(X)]\|_\infty>\epsilon) \leq 2d_q \exp(kd_q\log A - 2n\epsilon^2).$$
\end{lemma}

Now we show that contractiveness implied by the progressivenes condition forces the full analyst to be close in norm to its corresponding truncated version.

\begin{lemma}
\label{closeness1}
Let $a_t\in[0,1]^{d_q}$ for all $t\in\N$. For any $k\in\N$, the progressive analyst and the corresponding truncated analyst satisfy $\|h_t^k - h_t\|\leq \frac{\lambda^k L C_{\mathbf{1}}}{1-\lambda}$,
where $C_{\mathbf{1}}:=\|(1,\dots,1)\|$ is the norm of the $d_q$-dimensional all-ones vector.
\end{lemma}

\subsection{Generalization via Compression}
\label{gen1}

For a large enough level of truncation $k$, which depends on the radius $\Delta$ of the set of all possible histories $\HH$, the truncated analyst and the full analyst are identical. This level of truncation is time-independent, and hence, by Lemma \ref{boundedmem1}, progressive analysts also have a time-independent scaling of the generalization error.

\begin{theorem}
\label{generalize1}
Answering $t$ queries chosen adaptively by a $\lambda$-progressive analyst by rounding the empirical answer to $O(1/n)$ precision achieves overall generalization error at most $\tilde O(\sqrt{K(\lambda)d_q\log(t)/n})$, where $K(\lambda) = \frac{\log\left(\frac{LC_{\mathbf{1}}}{(1-\lambda)\lambda\Delta}\right)}{\log(1/\lambda)}$.
\end{theorem}

In other words, having $n= \tilde O(K(\lambda)d_q \log(t)/\epsilon^2)$ samples suffices to guarantee $\epsilon$-generalization error with high probability.

Let $\lambda = 1 - \frac{1}{t}$. Then, by the first-order Taylor approximation, $\log(1/\lambda) \approx \frac{1}{t}$, and hence the generalization error of Theorem \ref{generalize1} grows as $\tilde O(\sqrt{t d_q/n})$. The same scaling of generalization error is achieved by fully adaptive analysts in the case of $d_q$-dimensional queries, when there is no use of privacy mechanisms. As argued earlier, $\lambda=1$ corresponds to full adaptivity, so it comes as no surprise that the same rate is achieved.

Note also that the generalization error is completely independent of the dimension of the \emph{history} $d$. This justifies our ``infinite-dimensional'' example earlier in this section.


\subsection{Limitations}

\textbf{Differential privacy.}
It is natural to wonder why we never used differential privacy to prevent
progressive analysts from overfitting. In the proof of Theorem \ref{generalize1}, we allow the statistical mechanism to
return unobscured empirical answers (up to a small rounding error), although we initially argued that differentially private perturbations provide a quadratic improvement.

The main reason is that the truncated analyst \emph{does not} in general have a time-independent composition of differential privacy, in spite of the fact that the number of observed answers is time-independent. This follows from the observation that it receives answers of the full analyst, whose uncertainty grows with time. On a high level, changing one element in the data set $\Ss$ allows minor changes in the history in each step, even if differential privacy is used. After $t-k$ steps, for some $k\in\N$, these changes might pile up to lead to a completely different query than the one that resulted from the original data set $\Ss$. The initial input from $\Ss$ of the truncated analyst is the answer to this query, which is highly unstable for large enough $t$. Therefore, claiming time-independent generalization, if possible, would require a novel framework for designing mechanisms for adaptive data analysis, one that does not rely on differential privacy.

\textbf{Naive definition.}
Stepping away from the dynamical systems perspective for a moment, one might argue that a simple way to smoothly interpolate between no adaptivity and full adaptivity through recency bias is to truncate the analyst's view of the transcript. More formally, define $q^K_t = g^K_t(q^K_{t-1}, a^K_{t-1}, \dots, q^K_{t-K}, a^K_{t-K})$, for some fixed $K\in\N$ and functions $\{g^K_t\}$. The input to $g^K_t$ consists of the last $K$ query-response pairs. This seems to be in contrast with the usual adaptive query construction $q_t = g_t(q_{t-1}, a_{t-1}, \dots, q_1, a_1)$, for some $\{g_t\}$; here, the argument of $g_t$ is \emph{all} query-response pairs so far. However, we claim that this intuitive construction does not necessarily rule out full adaptivity.

\begin{claim}
\label{counterexample}
Suppose that an adaptive data analyst has a truncated view of the transcript with truncation depth $K$. In full generality, this analyst generalizes no better than an analyst with a full view of the transcript, regardless of the mechanism for constructing responses and value of $K$.
\end{claim}

\section{Conservative Analysts, Type A: Motivation and Generalization}
\label{conssec}

The second main class of natural analysts operates in a manner opposite to progressive analysts; namely,
these discount new evidence increasingly with time, making their
knowledge saturate. We will call such analysts \emph{conservative}.

We consider two possible causes for saturation. Either the maps $\{\psi_t\}$
become less sensitive to new evidence, or the queries $\{q_t\}$ are chosen in
such a way that the values $\{q_t(\Ss)\}$ saturate. This distinction leads to
two notions of conservative analysts. Type A conservatives and type B
conservatives. We will see that each correspond to natural algorithms.


Below we define type A conservative analysts, while we leave the definition of type B for the following section.

\begin{definition}[Conservative analyst, type A]
\label{analyst2}
An adaptive analyst is type A $\eta_t$-conservative if the maps $\{\psi_t(h,\cdot)\}$ are $\eta_t$-Lipschitz, where $\lim_{t\rightarrow\infty}\eta_t=0$. Mathematically, this corresponds to:
$$\|\psi_{t}(h,a) - \psi_{t}(h,a')\|\leq \eta_t \|a - a'\|,$$
for every $h\in\HH$ and $a,a'\in\A$.
\end{definition}

The construction of conservative analysts is primarily motivated by gradient descent in various settings in which it experiences saturation. As in the case of progressive analysts, however, there is also a connection between human data analysts and our definition of conservative analysts.

A common cognitive bias that humans experience in analytical tasks is called the anchoring bias \cite{campbell2009anchoring,cen2013role}. It is characterized by relying heavily on initial evidence, and becoming decreasingly sensitive to new evidence, as mathematically formulated in Definition \ref{analyst2}. The sequence $\{\eta_t\}$ in the definition of conservative analysts can be thought of as the strength of one's anchoring phenomenon. In particular, $\eta_t=0$ for all $t\in\N$ implies complete anchoring and no adaptivity in formulating queries, while a slow decrease in $\eta_t$ represents analysts with a mild anchoring effect.

From the algorithmic perspective, examples of conservative analysts include optimization algorithms with decaying step size. Consider the problem of empirical risk minimization using gradient descent. In particular, let the loss be:
$$L(h) = \frac{1}{n}\sum_{i=1}^n \ell(h;X_i),$$
where $h\in\R^d$ is a vector of weights for the given optimization model, and $\ell(h;X_i)$ is the loss incurred by this model on sample $X_i\in\Ss$. The well-known gradient descent update is the following:
\begin{equation*}
h_{t+1}=\psi_{t+1}(h_t, \nabla_h L(h_t)) = h_t-\eta_t \nabla_h L(h_t),
\end{equation*}
where $\nabla_h L(h_t)$ is the gradient of the loss on data $\Ss$ at point $h_t$, and $\eta_t$ is a time-dependent, decreasing step. Notice that this gradient decomposes as:
$$\nabla_h L(h_t) = \frac{1}{n}\sum_{i=1}^n \nabla_h \ell(h_t;X_i).$$
Therefore, gradient descent for empirical risk minimization is an $\eta_t$-conservative analyst, whose queries are equal to the gradient of the loss incurred at each point of $\Ss$ at the current weight iterate.

The rate of step size decay determines the rate of saturation of the analyst, allowing the class of conservative analysts to cover a wide spectrum of gradient-based optimization algorithms. Notable examples of step size decays include $\eta_t=O(1/t^\alpha)$, where $\alpha\in(0.5,1]$ \cite{robbins1951stochastic}, or the more recent schemes which cut the learning rate by a constant factor in every so-called epoch, which implies an essentially exponential decay \cite{hazan2014beyond, ge2018rethinking}.

\subsection{Truncated Analyst}

As its name suggests, the adaptiveness of a conservative analyst essentially saturates after some number of rounds of interaction with the data set. Again, we prove this via truncation of the full analyst. Let the truncated analyst corresponding to the full conservative analyst be the following:
\begin{align*}
    h_{t}^k &= \psi_t(h_{t-1}^k,a^k_{t-1}),\\
    \text{where}~~ a_t^k &= a_t, \forall t\leq k \text{ and } a_t^k=0, \forall t>k, \\
    q_t^k &= f_t(h_t^k).
\end{align*}
In words, the truncated analyst only sees the first $k$ true answers and deterministically sets the second input to 0 for the remaining $t-k$ rounds.

\begin{lemma}
\label{closeness2}
Assume that the answers of the statistical mechanism are bounded to $[0,1]^{d_q}$, and let $C_{\mathbf{1}}:=\|(1,\dots,1)\|$ denote the norm of the $d_q$-dimensional all-ones vector. Then, for $K(\eta_t):=\min\{t:\eta_t < \frac{\Delta}{C_{\mathbf{1}}}\}$, the history of the full analyst matches the history of the truncated analyst with truncation depth $K(\eta_t)$, $h_t = h_t^{K(\eta_t)}$.
\end{lemma}

Since the approximating truncated analyst in this setting sees the first $k$ answers, instead of the last $k$, the privacy parameters of its history degrade gracefully with $k$, given that the statistical mechanism is differentially private. The Gaussian mechanism is, however, not bounded, as required by Lemma \ref{closeness2}. For this reason, we introduce a slight modification of this mechanism, where the answers are computed as $a_t = [q_t(\Ss) + \xi_t]_{[0,1]^{d_q}}$, where $[\cdot]_{[0,1]^{d_q}}$ is truncation to the box $[0,1]^{d_q}$:
\begin{equation*}
  ([x]_{[0,1]^{d_q}})_i = \begin{cases}
        0, \text{ if } x_i\leq 0,\\
         x_i, \text{ if } x_i\leq 0,\\
		1, \text{ if } x_i\geq1,\\
        \end{cases}
 \end{equation*}
 where subscript $i$ denotes the $i$-th coordinate. As before, $\xi_t$ is $d_q$-dimensional Gaussian noise. By post-processing of differential privacy, this truncated mechanism preserves the parameters of differential privacy of the Gaussian mechanism, determined by the variance of $\xi_t$. 

The next lemma formalizes the gradual, time-independent degradation of differential privacy of the truncated analyst's history.

\begin{lemma}
\label{truncateddp2}
Let $h_t^k$ be the history of the truncated analyst at time $t\in\N$, and let the statistical mechanism be $(\alpha,\beta)$-differentially private. Then, $h_t^k$ is $(\sqrt{2k\log(1/\beta')}\alpha + 2k\alpha^2,k\beta + \beta')$-differentially private.
\end{lemma}

\subsection{Generalization via Differential Privacy}

Since we proved that type A conservative analysts have the same history as their corresponding truncated analyst, for a large enough level of truncation, and that truncated analysts have a time-independent composition of differential privacy, we can conclude a time-independent composition of privacy for the full analyst as well.

\begin{proposition}
\label{dpbound2}
Let $h_t$ be the hidden state of an oblivious analyst at time $t$. Let the statistical mechanism answering queries be $(\alpha,\beta)$-differentially private. Then, for \emph{arbitrarily large} $t$, $h_t$ is $(\sqrt{2K(\eta_t)\log(1/\beta')}\alpha + 2K(\eta_t)\alpha^2,K(\eta_t)\beta + \beta')$-differentially private, where $K(\eta_t) := \min\{t:\eta_t\leq \frac{\Delta}{C_{\mathbf{1}}}\}$.
\end{proposition}

To prove the generalization error of conservative analysts, we turn to the main transfer theorem of \citet{bassily2016algorithmic}, which is the main technical tool used to establish the celebrated rate of $\tilde O((t d_q)^{1/4}/\sqrt{n})$. This transfer theorem will allow us to compute the generalization error by balancing out the sample accuracy and differential privacy parameters of the Gaussian mechanism.

\begin{theorem}
\label{generalize2}
There is a computationally efficient mechanism to answer $t$ queries chosen adaptively by a type A $\eta_t$-conservative analyst so that the overall generalization error is at most $\tilde O((K(\eta_t) d_q \log(t))^{1/4}/\sqrt{n})$, where $K(\eta_t) := \min\{t:\eta_t\leq \frac{\Delta}{C_{\mathbf{1}}}\}$ and $C_{\mathbf{1}}=\|(1,\dots,1)\|$ is the norm of the $d_q$-dimensional all-ones vector.
\end{theorem}

Said in terms of sample complexity, it suffices to have $n = \tilde O(\sqrt{K(\eta_t)d_q\log(t)}/\epsilon^2)$ samples for $\epsilon$-generalization error. Notice that, just like for progressive analysts, there is no direct dependence on the dimension of the history.

As an example, taking $\eta_t = \eta^t$, where $\eta = 1 - 1/t$, results in error rate $\tilde O((t d_q)^{1/4}/\sqrt{n})$ by the first-order Taylor approximation. As expected, this is the tight rate for fully adaptive queries under differential privacy.

\section{Conservative Analysts, Type B: Motivation and Generalization}

In this section we define and analyze linear analysts whose histories saturate despite non-decreasing sensitivity to revealed information about $\Ss$.

\begin{definition}[Conservative analyst, type B]
\label{analyst3}
An adaptive analyst is type B $\lambda$-conservative if, first, it contracts when given empirical answers:
$$\| \psi_t(h_{t-1},q_{t-1}(S)) - \psi_t(h_{t-1}',q_{t-1}'(S))\| \leq \lambda\|h_{t-1} - h_{t-1}'\|,$$
for some $\lambda\in[0,1]$, where $q_t = f_t(h_t)$ and $q_t' = f_t(h_t')$. Second, we require the analyst to be linear:
$$h_t = \psi_t(h_{t-1},a_{t-1}) = A_t h_{t-1} + B_t a_{t-1},$$
for some sequences $\{A_t\},\{B_t\}$, where $A_i\in\R^{d\times d}$, with $\|A_i\|_{\text{op}}\leq 1$, and $B_i\in\R^{d\times d_q}$ for all $i\in\N$.
\end{definition}

The motivation for type B conservative analysts comes from the observation that gradient-based methods sometimes saturate even if there is no step decay.

Consider again the problem of empirical risk minimization using gradient descent. In this setting, let the gradient descent update have a constant step size $\eta>0$:
\begin{equation*}
h_{t+1}=\psi_{t+1}(h_t, \nabla_h L(h_t)) = h_t-\eta \nabla_h L(h_t),
\end{equation*}
where $ \nabla_h L(h_t) = \frac{1}{n}\sum_{i=1}^n \ell(h_t;X_i)$ is again the gradient of the loss on data $\Ss$. If the loss is $\beta$-smooth and $\mu$-strongly convex, and the step size is $\eta\leq \frac{2}{\beta + \mu}$, then the gradient descent update is type B $\lambda$-conservative, where $\lambda= 1 - \frac{\eta\beta\mu}{\beta + \mu}$ \cite{hardt2015train}. If the objective is not strongly convex, however is still smooth, gradient descent is non-expansive, meaning it has contraction parameter $\lambda=1$. In that case, one can induce contractiveness in many ways; one is to add an $\ell_2$-regularizer to the objective, that is transform the loss into $L^{\text{reg}}(h) = L(h) + \frac{\mu}{2}\|h\|^2$, for some $\mu>\beta$.  

\subsection{Truncated Analyst}

As in the previous section, due to saturation of conservative analysts, we will define a truncated analyst that has access to $k$ responses of the statistical mechanism. In this case, however, the interaction happens in the \emph{last} $k$ rounds.

Suppose that the statistical mechanism is the usual Gaussian mechanism. Worth mentioning is that this time we deploy no truncation. Denote by $\xi_t$ the noise variable added to the empirical answer at time $t$. For fixed $k$, define the truncated analyst corresponding to a type B conservative analyst as:
\begin{align*}
    h_t^k &= \psi_t(h_{t-1}^k,a^k_{t-1}), ~~ h_{t-j}^k = 0, \forall j\geq k,\\
    \text{ where } a_t^k &= q_t^k(\Ss) + \xi_t,\\
    q_t^k &= f_t(h_t^k),
\end{align*}

In this setting, we assume that $\HH$ is a norm-ball with radius $D$, where $D$ is large enough with respect to $\sum_{i=1}^t \|B_i\|_{\text{op}}C_{\mathbf{1}}$, so that there is no need for projecting the norm of the current history iterates to $D$. Since this ``escaping'' event happens with negligible probability, in all subsequent arguments for simplicity we treat it as being of measure zero. First we establish closeness between the full analyst and the truncated version.

\begin{lemma}
\label{closeness3}
Suppose that the statistical mechanism is the Gaussian mechanism. For any $k\in\N$, the truncated analyst with truncation depth $k$ and the full analyst satisfy $\|h_t - h_t^k\| \leq \lambda^k D$.
\end{lemma}

Additionally, we show that the truncated analyst has a composition of differential privacy which only depends on the truncation depth.

\begin{lemma}
\label{boundedmem3}
Let $h_t^k$ be the history of a truncated analyst corresponding to a type B conservative analyst, and let the statistical mechanism be $(\alpha,\beta)$-differentially private. Then, for all $t\in\N$ and $\beta'>0$, $h_t^k$ is $(\sqrt{2k\log(1/\beta')}\alpha + 2k\alpha^2,k\beta + \beta')$-differentially private.
\end{lemma}

\subsection{Generalization via Differential Privacy}

Lemma \ref{closeness3} allows us to find the effective memory of a conservative analyst, resulting in the following time-independent composition of differential privacy parameters.

\begin{proposition}
\label{dpbound3}
Let $h_t$ be the history of a type B conservative analyst at time $t$. Let the statistical mechanism answering queries be the Gaussian mechanism, such that the answers are $(\alpha,\beta)$-differentially private. Then, for arbitrarily large $t$, $h_t$ is $(\sqrt{2K(\lambda)\log(1/\beta')}\alpha + 2K(\lambda)\alpha^2,K(\lambda)\beta + \beta')$-differentially private, where $K(\lambda) := \frac{\log(D/\Delta\lambda)}{\log(1/\lambda)}$.
\end{proposition}

The main transfer theorem of \citet{bassily2016algorithmic} will now show that the generalization error of type B conservative analysts is essentially the same as for type A conservative analysts, justifying their unification into one broader class.

\begin{theorem}
\label{generalize3}
There is a computationally efficient mechanism to answer $t$ queries chosen adaptively by a type B $\lambda$-conservative analyst so that the overall generalization error is at most $\tilde O((K(\lambda) d_q \log(t))^{1/4}/\sqrt{n})$, where $K(\lambda) := \frac{\log(D/\Delta\lambda)}{\log(1/\lambda)}$.
\end{theorem}

In other words, $n = \tilde O(\sqrt{K(\lambda)d_q\log(t)}/\epsilon^2)$ samples suffice for $\epsilon$-generalization error. Moreover, under a few additional commonly satisfied assumptions, this sample complexity holds for much more general sets $\HH$, which need not have a finite resolution. This essentially follows by linearity of maps $\psi_t$. We prove this claim in the following section.

\section{Beyond the Discrete Setting}
\label{beyond}

In this section, we drop the assumption that $\HH$ is discrete with $\Delta$-resolution. First we prove that, under no additional regularity, progressive and type A conservative analysts can be arbitrarily adaptive, regardless of their parameters of contraction. We do, however, eliminate the trivial cases: $\lambda=0$ or $L=0$ in Definition \ref{analyst1}, and $\eta_t = 0$ in Definition \ref{analyst2}.

\begin{proposition}
\label{negative}
Without any assumption on $\HH$, progressive and type A conservative analysts can overfit as much as an adaptive analyst with a full view of the transcript, as long as the parameters of contraction are non-trivial.
\end{proposition}

The previous proposition shows that general contractive maps $\psi_t$ do not ensure better generalization in continuous settings. Under mild regularity, however, linear maps imply that adding noise to the truthful answer is essentially the same as adding noise to the history of the analyst. We exploit this observation in the following theorem, proving a result quantitatively almost identical to that of Theorem \ref{generalize3}, although for more general sets $\HH$.

\begin{theorem}
\label{linearcts}
Let $d_q=d$, and suppose $\{B_t\}$ is a sequence of positive-definite or negative-definite matrices, where $\lambda_{\min}:=\min_t \min_i |\lambda_i (B_t)|$. Then, without any discretization assumption on $\HH$,  there is a computationally efficient mechanism to accurately answers $t$ queries chosen adaptively by a type B $\lambda$-conservative analyst, given $n = \tilde O(\sqrt{K(\lambda) d \log(t)}/\epsilon^2)$ samples, where $K(\lambda) = O\left(\frac{\log(D/\sqrt{\lambda_{\min}})}{\log(1/\lambda)}\right)$.
\end{theorem}


\section{Summary}

We introduce progressive and conservative analysts by modeling the evolution of
their knowledge using different control-theoretic constraints. In addition to
serving as mathematical analogies of human cognitive biases, these categories also capture various iterative algorithms, like value iteration or gradient descent. 
The natural analysts we define achieve generalization error essentially
independent of the number of queries, in stark contrast with arbitrary
adversarial analysts whose error scales polynomially. In doing so, we combine
control-theoretic notions of stability with the algorithmic stability notions underpinning
adaptive generalization bounds. The connection between control-theoretic and
algorithmic stability for the sake of proving stronger generalization bounds is
worth studying further.

\section*{Acknowledgements}

The authors thank John Miller for his careful reading of a draft of this paper and constructive feedback provided.


\begin{thebibliography}{21}
\providecommand{\natexlab}[1]{#1}
\providecommand{\url}[1]{\texttt{#1}}
\expandafter\ifx\csname urlstyle\endcsname\relax
  \providecommand{\doi}[1]{doi: #1}\else
  \providecommand{\doi}{doi: \begingroup \urlstyle{rm}\Url}\fi

\bibitem[Bassily et~al.(2016)Bassily, Nissim, Smith, Steinke, Stemmer, and
  Ullman]{bassily2016algorithmic}
Raef Bassily, Kobbi Nissim, Adam Smith, Thomas Steinke, Uri Stemmer, and
  Jonathan Ullman.
\newblock Algorithmic stability for adaptive data analysis.
\newblock In \emph{Proceedings of the forty-eighth annual ACM symposium on
  Theory of Computing}, pages 1046--1059. ACM, 2016.

\bibitem[Bellman(1957)]{bellman1957dynamic}
Richard~Ernest Bellman.
\newblock Dynamic programming.
\newblock \emph{Princeton University Press}, 1957.

\bibitem[Belloni et~al.(2014)Belloni, Chernozhukov, and
  Hansen]{belloni2014inference}
Alexandre Belloni, Victor Chernozhukov, and Christian Hansen.
\newblock Inference on treatment effects after selection among high-dimensional
  controls.
\newblock \emph{The Review of Economic Studies}, 81\penalty0 (2):\penalty0
  608--650, 2014.

\bibitem[Bertsekas(2005)]{bertsekas2005dynamic}
Dimitri~P Bertsekas.
\newblock \emph{Dynamic programming and optimal control}, volume~1.
\newblock Athena scientific Belmont, MA, 2005.

\bibitem[Blum and Hardt(2015)]{blum2015ladder}
Avrim Blum and Moritz Hardt.
\newblock The ladder: A reliable leaderboard for machine learning competitions.
\newblock In \emph{International Conference on Machine Learning}, pages
  1006--1014, 2015.

\bibitem[Boucheron et~al.(2013)Boucheron, Lugosi, and
  Massart]{boucheron2013concentration}
St{\'e}phane Boucheron, G{\'a}bor Lugosi, and Pascal Massart.
\newblock \emph{Concentration inequalities: A nonasymptotic theory of
  independence}.
\newblock Oxford university press, 2013.

\bibitem[Campbell and Sharpe(2009)]{campbell2009anchoring}
Sean~D Campbell and Steven~A Sharpe.
\newblock Anchoring bias in consensus forecasts and its effect on market
  prices.
\newblock \emph{Journal of Financial and Quantitative Analysis}, 44\penalty0
  (2):\penalty0 369--390, 2009.

\bibitem[Cen et~al.(2013)Cen, Hilary, and Wei]{cen2013role}
Ling Cen, Gilles Hilary, and KC~John Wei.
\newblock The role of anchoring bias in the equity market: Evidence from
  analysts' earnings forecasts and stock returns.
\newblock \emph{Journal of Financial and Quantitative Analysis}, 48\penalty0
  (1):\penalty0 47--76, 2013.

\bibitem[Cheadle et~al.(2014)Cheadle, Wyart, Tsetsos, Myers, De~Gardelle,
  Casta{\~n}{\'o}n, and Summerfield]{cheadle2014adaptive}
Samuel Cheadle, Valentin Wyart, Konstantinos Tsetsos, Nicholas Myers, Vincent
  De~Gardelle, Santiago~Herce Casta{\~n}{\'o}n, and Christopher Summerfield.
\newblock Adaptive gain control during human perceptual choice.
\newblock \emph{Neuron}, 81\penalty0 (6):\penalty0 1429--1441, 2014.

\bibitem[Dwork and Roth(2014)]{dwork2014algorithmic}
Cynthia Dwork and Aaron Roth.
\newblock The algorithmic foundations of differential privacy.
\newblock \emph{Foundations and Trends{\textregistered} in Theoretical Computer
  Science}, 9\penalty0 (3--4):\penalty0 211--407, 2014.

\bibitem[Dwork et~al.(2010)Dwork, Rothblum, and Vadhan]{dwork2010boosting}
Cynthia Dwork, Guy~N Rothblum, and Salil Vadhan.
\newblock Boosting and differential privacy.
\newblock In \emph{2010 IEEE 51st Annual Symposium on Foundations of Computer
  Science}, pages 51--60. IEEE, 2010.

\bibitem[Dwork et~al.(2015{\natexlab{a}})Dwork, Feldman, Hardt, Pitassi,
  Reingold, and Roth]{dwork2015generalization}
Cynthia Dwork, Vitaly Feldman, Moritz Hardt, Toni Pitassi, Omer Reingold, and
  Aaron Roth.
\newblock Generalization in adaptive data analysis and holdout reuse.
\newblock In \emph{Advances in Neural Information Processing Systems}, pages
  2350--2358, 2015{\natexlab{a}}.

\bibitem[Dwork et~al.(2015{\natexlab{b}})Dwork, Feldman, Hardt, Pitassi,
  Reingold, and Roth]{dwork2015preserving}
Cynthia Dwork, Vitaly Feldman, Moritz Hardt, Toniann Pitassi, Omer Reingold,
  and Aaron~Leon Roth.
\newblock Preserving statistical validity in adaptive data analysis.
\newblock In \emph{Proceedings of the forty-seventh annual ACM symposium on
  Theory of computing}, pages 117--126. ACM, 2015{\natexlab{b}}.

\bibitem[Fithian et~al.(2014)Fithian, Sun, and Taylor]{fithian2014optimal}
William Fithian, Dennis Sun, and Jonathan Taylor.
\newblock Optimal inference after model selection.
\newblock \emph{arXiv preprint arXiv:1410.2597}, 2014.

\bibitem[Ge et~al.(2018)Ge, Kakade, Kidambi, and Netrapalli]{ge2018rethinking}
Rong Ge, Sham~M Kakade, Rahul Kidambi, and Praneeth Netrapalli.
\newblock Rethinking learning rate schedules for stochastic optimization.
\newblock 2018.

\bibitem[Hardt and Ullman(2014)]{hardt2014preventing}
Moritz Hardt and Jonathan Ullman.
\newblock Preventing false discovery in interactive data analysis is hard.
\newblock In \emph{Foundations of Computer Science (FOCS), 2014 IEEE 55th
  Annual Symposium on}, pages 454--463. IEEE, 2014.

\bibitem[Hardt et~al.(2016)Hardt, Recht, and Singer]{hardt2015train}
Moritz Hardt, Ben Recht, and Yoram Singer.
\newblock Train faster, generalize better: Stability of stochastic gradient
  descent.
\newblock In \emph{International Conference on Machine Learning}, pages
  1225--1234, 2016.

\bibitem[Hazan and Kale(2014)]{hazan2014beyond}
Elad Hazan and Satyen Kale.
\newblock Beyond the regret minimization barrier: optimal algorithms for
  stochastic strongly-convex optimization.
\newblock \emph{The Journal of Machine Learning Research}, 15\penalty0
  (1):\penalty0 2489--2512, 2014.

\bibitem[Miller and Hardt(2019)]{miller2018recurrent}
John Miller and Moritz Hardt.
\newblock Stable recurrent models.
\newblock In \emph{International Conference on Learning Representations (to
  appear)}, 2019.

\bibitem[Robbins and Monro(1951)]{robbins1951stochastic}
Herbert Robbins and Sutton Monro.
\newblock A stochastic approximation method.
\newblock \emph{The Annals of Mathematical Statistics}, 22\penalty0
  (3):\penalty0 400--407, 1951.

\bibitem[Ullman et~al.(2018)Ullman, Smith, Nissim, Stemmer, and
  Steinke]{ullman2018limits}
Jonathan Ullman, Adam Smith, Kobbi Nissim, Uri Stemmer, and Thomas Steinke.
\newblock The limits of post-selection generalization.
\newblock In \emph{Advances in Neural Information Processing Systems}, pages
  6402--6411, 2018.

\end{thebibliography}

\section{Preliminaries on Differential Privacy}

Here we review two useful properties of differential privacy, which are repeatedly utilized throughout this manuscript.  The omitted proofs can be found in the book \cite{dwork2014algorithmic}.

The first lemma states that the output of a differentially private algorithm remains private under any subsequent post-processing.



\begin{lemma}[Post-processing of differential privacy]
\label{postprocessing}
Let $\F_2:\X\rightarrow \Y$ be an arbitrary randomized function, and let $\F_1:\D^n\rightarrow \X$ be an $(\alpha,\beta)$-differentially private mapping. Then, $\F_2 \circ \F_1$ is also $(\alpha,\beta)$-differentially private.
\end{lemma}

Differential privacy also has favorable composition properties. Linear composition is easy to show by definition of differential privacy, while for the quadratic improvement stated in Lemma \ref{compositiondp} one requires more sophisticated proof techniques, originally outlined in the paper \cite{dwork2010boosting}.

\begin{lemma}[Linear composition of differential privacy]
\label{lincompositiondp}
Let $\F_1:\D^n\rightarrow\Y$ be an $(\alpha_1,\beta_1)$-differentially private mapping of a data set $\Ss$, and let $\F_2:\Y\times\D^n\rightarrow\Y$ be $(\alpha_2,\beta_2)$-differentially private for every fixed $y_1\in\Y$. Then, the composition of $\F_1$ and $\F_2$, obtained as $\F_2(\F_{1}(\Ss))$, is  $(\alpha_1 + \alpha_2,\beta_1 + \beta_2)$-differentially private.
\end{lemma}

\begin{lemma}[Strong composition of differential privacy]
\label{compositiondp}
Let $\F_1:\D^n\rightarrow\Y$ be an $(\alpha,\beta)$-differentially private mapping of a data set $\Ss$, and for every $i\geq 2$, let $\F_i:\Y^{i-1}\times \D^n\rightarrow \Y$ be $(\alpha,\beta)$-differentially private for every fixed $y_1,\dots,y_{i-1}\in\Y^{i-1}$. Then, the composition of $\F_1,\dots,\F_i$ obtained as $\F_i(\F_{i-1}(\dots,\Ss)\Ss)$ is $(\sqrt{2i\log(1/\beta')}\alpha + i\alpha(e^\alpha-1),i\beta + \beta')$-differentially private, for any $\beta'\in(0,1]$.
\end{lemma}

Since $e^\alpha-1 \leq 2\alpha$ for $\alpha\in[0,1]$, this paper uses a more convenient composition bound, which states that an $i$-fold composition of $(\alpha,\beta)$-differentially private algorithms is $(\sqrt{2i\log(1/\beta')}\alpha + 2i\alpha^2,i\beta + \beta')$-differentially private.

This work mainly focuses on the Gaussian mechanism for achieving differential privacy, hence we review the privacy properties of this method.

\begin{lemma}[Properties of the Gaussian mechanism] Take any $u,v\in\R^{d_q}$, and let $\xi_1,\xi_2\sim N(0,\sigma^2 I_{d_q})$ be two independent $d_q$-dimensional Gaussian noise vectors. Denote $u^\xi = u + \xi_1$ and $v^\xi = v + \xi_2$. Then, it holds that:
\begin{align*}
    \PP(u^\xi\in\Oh)\leq &\exp\left(\frac{\sqrt{2\log(1.25/\beta)}\|u-v\|_2}{\sigma}\right)\PP(v^\xi\in\Oh) + \beta,
\end{align*}
for any $\beta>0$.
\end{lemma}

If $u$ and $v$ represent $d_q$-dimensional empirical answers to the same query on data sets $S$ and $S'$, respectively, where $S$ and $S'$ differ in at most one element, we have $\|u-v\|_2\leq \frac{\sqrt{d_q}}{n}$. As a result, the Gaussian mechanism with parameter $\sigma$ is $(\sqrt{2\log(1.25/\beta)}\sqrt{d_q}/n\sigma,\beta)$-differentially private, for any $\beta>0$.




\section{Progressive Analysts: Proofs}

\subsection{Proof of Lemma \ref{boundedmem1}}

The claim follows by applying a union bound together with the Hoeffding concentration bound. In particular, for every \emph{fixed} $d_q$-dimensional statistical query $q$ and target accuracy $\epsilon$, the following is true by the Hoeffding bound:
$$\PP(\|q(\Ss) - \EE_{X\sim\Pp}[q(X)]\|_\infty >\epsilon)\leq 2d_q \exp(-2n\epsilon^2),$$
where we take a union bound over the coordinates of $q$. Now notice that, by definition of the truncated analyst, we can write $h_t^k$, and consequently also $q_t^k$, as a function of $a^k := (a_{t-k},\dots,a_{t-1})$. There are $A^{k d_q}$ possibilities for the value of $a^k$. With this, we can take the union bound over all possibilities for $q^k_t$ to conclude:
\begin{align*}
    \PP(\| q^k_t(\Ss) - \EE_{X\sim\Pp}[q^k_t(X)]\|_\infty > \epsilon) &\leq 2d_q A^{k d_q} \exp(-2n\epsilon^2)\\
    &= 2d_q \exp(kd_q\log A - 2n\epsilon^2).
\end{align*}
Since $A$ is polynomial in $n$, we have that the generalization error scales as $\tilde O(\sqrt{kd_q/n})$.

\subsection{Proof of Lemma \ref{closeness1}}

The claimed bound is a consequence of the definition of progressiveness. First, because the truncated and full analyst receive the same answers, we have:
\begin{align*}
    \|h_t - h_t^k\|&=\| \psi_t(h_{t-1},a_{t-1}) - \psi_t(h^k_{t-1},a_{t-1}) \| \\
    &\leq\lambda\|h_{t-1}-h_{t-1}^k\|\\
    &\leq\lambda^k\|h_{t-k}\|,
\end{align*}
where the last step follows because $h_{t-k}^k=0$. Now we exploit the Lipschitz properties of the maps $\{\psi_t \}$:
\begin{align*}
     \|h_t - h_t^k\|&\leq \lambda^k\|\psi_t(h_{t-k-1},a_{t-k-1}) - \psi_t(0,0)\|\\
    &= \lambda^k\|\psi_t(h_{t-k-1},a_{t-k-1}) - \psi_t(0,0) + \psi_t(0,a_{t-k-1}) - \psi_t(0,a_{t-k-1})\|\\
    &\leq \lambda^k(\|\psi_t(h_{t-k-1},a_{t-k-1}) - \psi_t(0,a_{t-k-1})\| + \|\psi_t(0,a_{t-k-1}) - \psi_t(0,0)\|)\\
    &\leq \lambda^k (\lambda \|h_{t-k-1}\| + L\|a_{t-k-1}\|)\\
    &\leq \frac{\lambda^k LC_{\mathbf{1}}}{1-\lambda},
\end{align*}
where the last step follows by recursively applying the same steps to the term $\|h_{t-k-1}\|$, and due to the fact that $\|a_{t-k-1}\|\leq \|(1,\dots,1)\|$.

\subsection{Proof of Theorem \ref{generalize1}}

Take any $h,h'\in\HH$, such that $h\neq h'$. Then, $h$ and $h'$ have to differ by at least $\Delta$ in norm, assuming that they are equal in all coordinates but one. However, by Lemma \ref{closeness1}, we have:
$$\|h_t - h_t^k\|\leq \frac{\lambda^k LC_{\mathbf{1}}}{1-\lambda}.$$
This means that, if $\frac{\lambda^k LC_{\mathbf{1}}}{1-\lambda}< \Delta$, $h_t$ and $h_t^k$ are identical. In other words, a truncated analyst with truncation level $\left\lceil\frac{\log\left(\frac{LC_{\mathbf{1}}}{(1-\lambda)\Delta}\right)}{\log(1/\lambda)}\right\rceil\leq\frac{\log\left(\frac{LC_{\mathbf{1}}}{(1-\lambda)\Delta}\right)}{\log(1/\lambda)} + 1 = \frac{\log\left(\frac{LC_{\mathbf{1}}}{(1-\lambda)\lambda\Delta}\right)}{\log(1/\lambda)}:=K(\lambda)$ is identical to the corresponding progressive analyst. Since queries are determined solely by the value of the current history, the queries asked by the full analyst and its truncated version at time $t$ have to be identical. Let each answer be constructed as the projection of the empirical answer to the set $\A=\{0,\frac{\epsilon}{2n},\frac{\epsilon}{n},\dots,1\}^{d_q}$. Then, by a union bound:
\begin{align*}
    \PP(\max_{1\leq i \leq t}\| q_i(\Ss) - \EE_{X\sim\Pp}[q_i(X)]\|_\infty > \epsilon) &\leq \sum_{i=1}^t \PP(\| q_i(\Ss) - \EE_{X\sim\Pp}[q_i(X)]\|_\infty > \epsilon)\\
    &\leq 2td_q\exp(K(\lambda)d_q\log(2n/\epsilon+1) - 2n\epsilon^2),
\end{align*}
where the last step applies Lemma \ref{boundedmem1}. Since $\|q_i(\Ss) - a_i\|_\infty\leq O\left(\frac{1}{n}\right)$, we can conclude that the generalization error scales as $\tilde O(\sqrt{K(\lambda)d_q\log(t)/n})$.

\subsection{Proof of Claim \ref{counterexample}}

Suppose that the data samples are supported on $\R$, with no atoms. Let $d_q$ and $d_a$ be the dimensions of the queries and query results, respectively. Typically $d_q=d_a$, but this assumption is not necessary for the current counterexample. We will prove the claim for $K=1$, which will immediately imply the claim for all $K\in\N$. Let $g_1$ be any bijection between $\R^{d_a}$ and $\R$, and $g_2$ be any bijection between $\R^2$ and $\R$; the existence of such functions is a standard set-theoretic result. Pick a ``reserved value'' $r \in [0,1]^{d_q}$. All queries the analyst wishes to ask while interacting with the response mechanism must have the inverse image of $r$ to be a singleton; this does not effectively limit the scope of queries, since $r$ can have infinite precision. After the first round, set $q_2(g_1(a_1)) = r$. In all higher rounds $t\geq 3$, set $q_t(g_2(g_1(a_{t-1}), q_{t-1}^{-1}(r))) = r$. Since $g_1$ and $g_2$ are bijections, at any round $t$ one can recover $a_{t-1}$, as well as the previous encoding of the transcript $q_{t-1}^{-1}(r)$, which allows recursive recovery of all answers $a_1,a_2,\dots,a_{t-1}$. Since the queries are constructed deterministically based on the current transcript, knowing all answers encodes all query-answer pairs in a lossless fashion. Therefore, despite only having access to $q_{t-1}$ and $a_{t-1}$ at time $t$, the analyst is familiar with the full transcript. Consequently, the analyst can be arbitrarily adaptive, which completes the proof.

\section{Conservative Analysts, Type A: Proofs}

\subsection{Proof of Lemma \ref{closeness2}}

Fix $h_{t-1}$, and take two different answers $a_{t-1},a_{t-1}'\in\A$. Denote the histories resulting from evolving $h_{t-1}$ using $a_{t-1}$ and $a_{t-1}'$ by $h_t$ and $h_t'$, respectively. Then, by definition of type A conservative analysts:
\begin{align*}
    \|h_t - h_t'\|&= \|\psi_t(h_{t-1},a_{t-1}) - \psi_t(h_{t-1},a_{t-1}')\|\\
    &\leq \eta_t \|a_{t-1} - a_{t-1}'\|\\
    &\leq \eta_t C_{\mathbf{1}},
\end{align*}
where $C_{\mathbf{1}}:=\|(1,\dots,1)\|$, which follows from the assumption that the answers are bounded to $[0,1]^{d_q}$. Since the set $\HH$ has $\Delta$-resolution, if $h_t\neq h_t'$, it has to hold that $\|h_t- h_t'\|\geq \Delta$. Therefore, if $\eta_t C_{\mathbf{1}} < \Delta$, the history is determined solely depending on $h_{t-1}$ and $\psi_t$, with no dependence on $a_{t-1}$ and $a_{t-1}'$. Denote $K(\eta_t) := \min\{t: \eta_t C_{\mathbf{1}} < \Delta\}$. Since $\eta_t$ is a non-increasing sequence, using recursive reasoning one can conclude that the history at all times after $K(\eta_t)$ does not depend on the value of the current answer. As a result, we can set all answers after time $K(\eta_t)$ to be equal to 0, with no change on the analyst's history sequence. This exactly means that $h_t^{K(\eta_t)} = h_t$, for $K(\eta_t):=\{\min t: \eta_t < \Delta/C_{\mathbf{1}}\}$, which completes the proof.

\subsection{Proof of Lemma \ref{truncateddp2}}

By the strong composition of differential privacy, $h_k^k$ is $(\sqrt{2k\log(1/\beta')}\alpha + 2k\alpha^2,k\beta + \beta')$-differentially private. For all rounds after the $k$-th one, the answers are constant and independent of the data set $\Ss$, meaning they are $(0,0)$-differentially private. Hence, by the linear composition of differential privacy, for all $t\geq k$ the history remains $(\sqrt{2k\log(1/\beta')}\alpha + 2k\alpha^2,k\beta + \beta')$-differentially private. Both composition results are stated in the Appendix.

\subsection{Proof of Proposition \ref{dpbound2}}

The proof follows directly from Lemma \ref{closeness2} and Lemma \ref{truncateddp2}.

\subsection{Proof of Theorem \ref{generalize2}}

Let the statistical mechanism be the truncated Gaussian mechanism with parameter $\sigma=\frac{\epsilon}{\sqrt{2\log (2td_q/\epsilon\delta)}}$; that is, the answers are constructed as $a_t = [q_t(\Ss) + \xi_t]_{[0,1]^{d_q}}$, where $\xi_t$ is a $d_q$-dimensional vector with entries distributed as $N(0,\sigma^2)$. Then, by the sub-gaussian tail bound \cite{boucheron2013concentration}, as well as a union bound:
\begin{align*}
    \PP(\max_{1\leq i \leq t}\|a_i - q_i(\Ss)\|_\infty \geq \epsilon)&\leq \PP(\max_{1\leq i \leq t}\| \xi_i \|_\infty \geq \epsilon)\\
    &\leq 2t d_q \exp\left(-\frac{\epsilon^2}{2\sigma^2}\right)\\
    &= \epsilon \delta,
\end{align*}
where in the last step we plug in the choice of $\sigma$. Therefore, such a mechanism is $(\epsilon,\epsilon\delta)$-sample accurate. By properties of the Gaussian mechanism, this mechanism is also
$$\left(\frac{\sqrt{4d_q\log (2td_q/\epsilon\delta) \log(1.25 K(\eta_t)/\epsilon\delta)}}{n\epsilon},\epsilon\delta/K(\eta_t)\right)$$
-differentially private. By Proposition \ref{dpbound2}, for an \emph{arbitrarily large} $t$, the history $h_t$ is
\small
\begin{align*}
    \Big(&\sqrt{2K(\eta_t)\log(1/\beta')}\frac{\sqrt{4d_q\log (2td_q/\epsilon\delta) \log(1.25K(\eta_t)/\epsilon\delta)}}{n\epsilon} + 2K(\eta_t)\frac{4d_q\log (2td_q/\epsilon\delta) \log(1.25K(\eta_t)/\epsilon\delta)}{n^2\epsilon^2},\epsilon\delta + \beta'\Big)
\end{align*}
\normalsize
-differentially private, for any $\beta'>0$. Given $n \geq \tilde O\left(\frac{\sqrt{K(\eta_t)d_q\log(t)}}{\epsilon^2}\right)$, this composition is $(O(\epsilon),O(\epsilon\delta))$-differentially private. By the main transfer theorem of \citet{bassily2016algorithmic}, having $(O(\epsilon), O(\epsilon\delta))$-differential privacy, as well as $(O(\epsilon), O(\epsilon\delta))$-sample accuracy, implies $\epsilon$-generalization error with probability at least $1-\delta$. Therefore, the generalization error scales as $\tilde O((K(\eta_t)d_q \log(t))^{1/4}/\sqrt{n})$.

\section{Conservative Analysts, Type B: Proofs}

\subsection{Proof of Lemma \ref{closeness3}}

To prove the claim, we exploit the fact that the truncated analyst and the full analyst receive the same noise variables at any given round. This implies:
\begin{align*}
    \|h_t - h_t^k\| &=\|A_t h_{t-1} + B_t q_{t-1}(\Ss) + B_t 
    \xi_{t-1} - A_t h_{t-1}^k - B_t q_{t-1} ^k(\Ss) - B_t 
    \xi_{t-1}\|\\
    &=\|\psi_t(h_{t-1},q_{t-1}(\Ss)) - \psi_t(h^k_{t-1},q^k_{t-1}(\Ss))\|\\
    &\leq \lambda\|h_{t-1} - h_{t-1}^k\|\\
    &\leq  \lambda^k \|h_{t-k}\|\\
    &\leq \lambda^k D,
\end{align*}
where the first equality follows by canceling the noise term, the first inequality uses the fact that the analyst is $\lambda$-conservative, the second inequality applies the previous argument recursively, and uses the fact that $h_{t-k}^k=0$, and the last inequality follows by the assumption of $\HH$ being bounded.

\subsection{Proof of Lemma \ref{boundedmem3}}

Since $h_{t-k}^k=0$, it is independent of the data set $\Ss$ and, as such, it must be $(0,0)$-differentially private. By linear composition of differential privacy, $h_{t-k+1}$ is then $(\alpha,\beta)$-differentially private. Moreover, in the last $k$ rounds, all indivudual answers are $(\alpha,\beta)$-differentially private, so by the strong composition of differential privacy, the history is $(\sqrt{2k\log(1/\beta')}\alpha + 2k\alpha^2,k\beta + \beta')$-differentially private. The composition results for differential privacy are stated in the Appendix.

\subsection{Proof of Proposition \ref{dpbound3}}

For any $h,h'\in\HH$, such that $h\neq h'$, it has to hold that $\|h-h'\|\geq\Delta$. However, as shown in Lemma \ref{closeness3}, for every $k$, $\|h_t^k - h_t\|\leq \lambda^k D$. Therefore, for truncation level $\Big\lceil \frac{\log(D/\Delta)}{\log(1/\lambda)}\Big\rceil\leq \frac{\log(D/\Delta\lambda)}{\log(1/\lambda)} := K(\lambda)$, $h_t^{K(\lambda)}=h_t$. Consequently, since $h_t^{K(\lambda)}$ is $(\sqrt{2K(\lambda)\log(1/\beta')}\alpha + 2K(\lambda)\alpha^2,K(\lambda)\beta + \beta')$-differentially private by Lemma \ref{boundedmem3}, then so is $h_t$.

\subsection{Proof of Theorem \ref{generalize3}}

Let the statistical mechanism be the Gaussian mechanism with parameter $\sigma=\frac{\epsilon}{\sqrt{2\log (2td_q/\epsilon\delta)}}$; that is, the answers are constructed as $a_t = q_t(\Ss) + \xi_t$, where $\xi_t$ is a $d_q$-dimensional vector with entries distributed as $N(0,\sigma^2)$. Then, by the sub-gaussian tail bound \cite{boucheron2013concentration}, as well as a union bound:
\begin{align*}
    \PP(\max_{1\leq i \leq t}\|a_i - q_i(\Ss)\|_\infty \geq \epsilon) &= \PP(\max_{1\leq i \leq t}\| \xi_i \|_\infty \geq \epsilon)\\
    &\leq 2t d_q \exp\left(-\frac{\epsilon^2}{2\sigma^2}\right)\\
    &= \epsilon \delta,
\end{align*}
where in the last step we plug in the choice of $\sigma$. Therefore, such a mechanism is $(\epsilon,\epsilon\delta)$-sample accurate. By properties of the Gaussian mechanism, this mechanism is also
$$\left(\frac{\sqrt{4d_q\log (2td_q/\epsilon\delta) \log(1.25 K(\lambda)/\epsilon\delta)}}{n\epsilon},\epsilon\delta/K(\lambda)\right)$$
-differentially private. By Proposition \ref{dpbound2}, for an \emph{arbitrarily large} $t$, the history $h_t$ is
\small
\begin{align*}
    \Big(&\sqrt{2K(\lambda)\log(1/\beta')}\frac{\sqrt{4d_q\log (2td_q/\epsilon\delta) \log(1.25K(\lambda)/\epsilon\delta)}}{n\epsilon} + 2K(\lambda)\frac{4d_q\log (2td_q/\epsilon\delta) \log(1.25K(\lambda)/\epsilon\delta)}{n^2\epsilon^2},\epsilon\delta + \beta'\Big)
\end{align*}
\normalsize
-differentially private, for any $\beta'>0$. Given $n \geq \tilde O\left(\frac{\sqrt{K(\lambda)d_q \log(t)}}{\epsilon^2}\right)$, this composition is $(O(\epsilon),O(\epsilon\delta))$-differentially private. By the main transfer theorem of \citet{bassily2016algorithmic}, having $(O(\epsilon), O(\epsilon\delta))$-differential privacy, as well as $(O(\epsilon), O(\epsilon\delta))$-sample accuracy, implies $\epsilon$-generalization error with probability at least $1-\delta$. Therefore, the generalization error scales as $\tilde O((K(\lambda)d_q \log(t))^{1/4}/\sqrt{n})$.

\section{Beyond the Discrete Setting: Proofs}

\subsection{Proof of Proposition \ref{negative}}

First we design a contraction that will be the main technical idea in the proof for both progressive and conservative analysts. Throughout we assume that answers and history are one-dimensional, however it is not hard to see that the idea easily extends to higher dimensions. In particular, the following argument should be applied coordinate-wise.

Suppose $h,a\in[0,1]$. If the real-valued answers happen to lie anywhere in $\R$, first contract them to $[0,1]$ using, for example, appropriately normalized arctangent.

Compute $h'$ by interlacing the decimals of $a$ and $h$. To be more precise, denote the decimals of $a$ as $a = 0.a^{1}a^{2}a^{3}\dots$. Similarly let $h = 0.h^{1}h^{2}h^{3}\dots$. Then, $h'$ is given by $h' = 0.a^{1}h^{1}a^{2}h^{2}\dots$; notice that this encoding allows perfect recovery of $a$ and $h$. Denote this construction by $h' = c(a,h)$.

Now we turn to progressive analysts. Fix any $\lambda\in(0,1)$ and $L>0$. Let $h_t = \psi_t(h_{t-1},a_{t-1}):= \min\{\lambda,L\} c(h_{t-1},a_{t-1})$ for all $t\in\N$. This mapping satisfies the conditions of Definition \ref{analyst1} and hence constitutes a $\lambda$-progressive analyst. However, $h_t$ and $a_t$ are compressed with no loss, allowing the whole transcript to be unrolled. Consequently, this analyst can be as adaptive as any data analyst with a full view of the transcript.

A similar argument proves the claim for conservative analysts. Suppose that the step sequence $\{\eta_t\}$ is an arbitrary non-increasing positive sequence. Let $h_t = \psi_t(h_{t-1},a_{t-1}):= \eta_t c(h_{t-1},a_{t-1})$ for all $t\in\N$. This update satisfies Definition \ref{analyst2} and, as such, represents a $\eta_t$-conservative analyst. As in the previous case, however, the hidden state is a lossless encoding of the transcript, and allows full adaptivity. 

This completes the proof of the proposition.

\subsection{Proof of Theorem \ref{linearcts}}

Fix a number of rounds $t\in\N$. Let the statistical mechanism be the Gaussian mechanism with parameter $\sigma=\frac{\epsilon}{\sqrt{2\log (2td/\epsilon\delta)}}$; that is, the answers are constructed as $a_t = q_t(\Ss) + \xi_t$, where $\xi_t$ is a $d$-dimensional vector with entries distributed as $N(0,\sigma^2)$. As stated in Theorem \ref{generalize3}, this mechanism is $(\epsilon,\epsilon\delta)$-sample accurate.

We can write $\xi_t = \xi_t^{(1)} + \xi_t^{(2)}$, where $\xi_t^{(1)},\xi_t^{(2)}\sim N(0,\sigma^2/2)$ are independent Gaussians. Further, we can rewrite the history update as:
$$h_t = \psi_t(h_{t-1}, q_{t-1}(\Ss) + \xi_{t-1}^{(1)}) + B_t \xi_t^{(2)} := h_t' + B_t \xi_{t-1}^{(2)},$$
where we exploit the linearity of the system. Now consider the variable:
$$h_t' =  \psi_t(h_{t-1}, q_{t-1}(\Ss) + \xi_{t-1}^{(1)}).$$
Let the truncated analyst $h_t^k$ corresponding to $h_t'$ be:
\begin{align*}
    h_t^k &= \psi_t(h_{t-1}^k,a^k_{t-1}), ~~ h_{t-j}^k = 0, \forall j\geq k,\\
    \text{ where } a_j^k &= q_j^k(\Ss) + \xi_j, \forall j<t, ~~ a_t^k = q_t^k(\Ss) + \xi^{(1)}_t,\\
    q_t^k &= f_t(h_t^k).
\end{align*}

In all rounds the truncated analyst gets noise with variance at least $\sigma^2/2$, so by Lemma \ref{boundedmem3}, as well as the properties of the Gaussian mechanism, $h_t^k$ is
$$\left(O(\sqrt{k d \log(1/\beta') \log (2td/\epsilon\delta)\log(1.25/\beta)}/\epsilon n),k\beta + \beta'\right)$$
-differentially private, for any $\beta,\beta'>0$. 
Recall that $h_t = h_t' + B_t \xi_{t-1}^{(2)}$, and take $h_t^{k,\xi} = h_t^k + B_t \xi^k_{t-1}$, where $\xi^k_{t-1}$ is an independent noise sample identically distributed as $\xi_{t-1}^{(2)}$.

Now we need to compute the parameters of differential privacy of $h_t$. Since stronger differential privacy only implies better generalization, we consider ``simplified'' versions of $h_t$ and $h_t^{k,\xi}$, which have less additive noise. One simplification would be to add a noise vector with independent entries, which are distributed as $N(0,\lambda_{\min}^2\sigma^2)$. The reason why this argument works is the following. First, zero-out all non-diagonal entries of $B_t$; by post-processing, this can only induce weaker differential privacy. Now notice that the diagonal entries of positive-definite or negative-definite matrices are in absolute value lower bounded by $\lambda_{\min}$. Therefore, setting the diagonal entries of $B_t$ to $\lambda_{\min}$ would again worsen the privacy parameters. In conclusion, the privacy parameters of $h_t$ and $h_t^{k,\xi}$ have to be at least as small as those of the simplified histories that would be obtained by adding a noise vector with independent entries that are distributed as $N(0,\lambda_{\min}^2\sigma^2)$, instead of $B_t \xi_{t-1}^{(2)}$ and $B_t \xi_{t-1}^k$, respectively.

By an analogous argument as in Lemma \ref{closeness3}, $\|h_t' - h_t^{k}\| \leq \lambda^k D$. To utilize the Gaussian mechanism, we need to bound $\|h_t' - h_t^{k}\|_2$. Since $\ell_p$-norms are decreasing in $p$, for $p\geq 1$, we can conclude that $\|h_t' - h_t^{k}\|_2 \leq \lambda^k D \sqrt{d}$, which follows by assuming contraction happens in $\ell_1$-norm, and, subsequently, by applying the Cauchy-Schwarz inequality.

Denote by $\nu_1,\nu_2$ two independent $d$-dimensional vectors whose entries are independent and distributed as $N(0,\lambda_{\min}^2\sigma^2)$. Define $h_{t,s} := h_t' + \nu_1$ and $h_{t,s}^{k,\xi} := h_t^k + \nu_2$. Take the depth of truncation to be $K(\lambda) = O\left(\frac{\log(D\sqrt{d}/\sqrt{\lambda_{\min}}\epsilon)}{\log(1/\lambda)}\right)$. Then, the properties of the Gaussian mechanism imply that $h_{t,s}$ and $h_{t,s}^{k,\xi}$ are indistinguishable; that is, for some constants $\epsilon'$ and $\delta'$:
\begin{equation}
\label{eq:indist1}
    \PP(h_{t,s}\in\Oh|\Ss = S)\leq e^{\epsilon'}\PP(h_{t,s}^{k,\xi}\in\Oh|\Ss = S) + \delta',
\end{equation}
and similarly:
\begin{equation}
\label{eq:indist2}
\PP(h_{t,s}^{k,\xi}\in\Oh|\Ss = S)\leq e^{\epsilon'}\PP(h_{t,s}\in\Oh|\Ss = S) + \delta'.
\end{equation}
By post-processing of differential privacy, we also know that $h_{t,s}^{k,\xi}$ has privacy parameters as least as good as $h_{t,s}^k$; to restate for convenience, $h_{t,s}^{k,\xi}$ is $\alpha^\xi,\beta^\xi)$-differentially private, where:
\begin{align*}
    \alpha^\xi&:=O(\sqrt{K(\lambda) d \log(1/\beta') \log (2td/\epsilon\delta)\log(1.25/\beta)}/\epsilon n),\\
    \beta^\xi&:=K(\lambda)\beta + \beta'.
\end{align*}
Recall that there exist constant parameters $\epsilon'$ and $\delta'$ such that equations \eqref{eq:indist1} and \eqref{eq:indist2} hold. Therefore, we have:
\begin{align*}
	\PP(h_{t,s} \in \Oh| \Ss = S)&\leq \exp(\epsilon') \PP(h_{t,s}^{k,\xi} \in \Oh | \Ss = S) + \delta'\\
	&\leq \exp(\epsilon' + \alpha^\xi) \PP(h_{t,s}^{k,\xi} \in \Oh | \Ss = S') + \exp\left(\epsilon'\right)\beta^\xi + \delta'\\
	&\leq \exp(2\epsilon' + \alpha^\xi) \PP(h_{t,s} \in \Oh | \Ss = S') + \exp(\epsilon' + \alpha^\xi)\delta'+ \exp(\epsilon')\beta^\xi + \delta'.
\end{align*}
To guarantee $(O(\epsilon),O(\epsilon\delta))$-differential privacy of $h_{t,s}$, the main requirement is to keep $\alpha^\xi$ proportional to $\epsilon$, as all other parameters can be chosen as arbitrarily small constants. This is achieved by having $n = \tilde O\left(\frac{\sqrt{K(\lambda) d\log(t)}}{\epsilon^2}\right)$ samples. Recall the main transfer theorem of \citet{bassily2016algorithmic}: having $(O(\epsilon), O(\epsilon\delta))$-differential privacy together with $(O(\epsilon), O(\epsilon\delta))$-sample accuracy implies $\epsilon$-generalization error with probability at least $1-\delta$. Applying this result completes the proof.

\end{document}